\documentclass[conference]{IEEEtran}
\usepackage{cite}
\usepackage{amsmath,amssymb,amsfonts}
\usepackage{algorithmic}
\usepackage{hyperref}
\usepackage{graphicx}
\usepackage{textcomp}
\usepackage{xcolor}
\def\BibTeX{{\rm B\kern-.05em{\sc i\kern-.025em b}\kern-.08em
    T\kern-.1667em\lower.7ex\hbox{E}\kern-.125emX}}

\usepackage{fancyhdr}
\begin{document}

\title{Socio-Emotional Response Generation: A Human Evaluation Protocol for LLM-Based Conversational Systems}

\author{\IEEEauthorblockN{Lorraine Vanel}
\IEEEauthorblockA{\textit{Zaion Lab, LTCI, ALMAnaCH} \\
\textit{Zaion, Télécom Paris, Inria}\\
Paris, France \\
lorraine.vanel@telecom-paris.fr}
\and
\IEEEauthorblockN{Ariel Ricardo Ramos Vela}
\IEEEauthorblockA{\textit{LTCI} \\
\textit{Télécom Paris}\\
Palaiseau, France \\
ariel.ram97@gmail.com}
\and
\IEEEauthorblockN{Alya Yacoubi}
\IEEEauthorblockA{\textit{Zaion Lab} \\
\textit{Zaion}\\
Paris, France \\
ayacoubi@zaion.ai}
\and
\IEEEauthorblockN{Chloé Clavel}
\IEEEauthorblockA{\textit{ALMAnaCH} \\
\textit{Inria}\\
Paris, France \\
chloe.clavel@inria.fr}
}

\maketitle

\begin{abstract}
Conversational systems are now capable of producing impressive and generally relevant responses. However, we have no visibility nor control of the socio-emotional strategies behind state-of-the-art Large Language Models (LLMs), which poses a problem in terms of their transparency and thus their trustworthiness for critical applications. Another issue is that current automated metrics are not able to properly evaluate the quality of generated responses beyond the dataset's ground truth. In this paper, we propose a neural architecture that includes an intermediate step in planning socio-emotional strategies before response generation. We compare the performance of open-source baseline LLMs to the outputs of these same models augmented with our planning module. We also contrast the outputs obtained from automated metrics and evaluation results provided by human annotators. We describe a novel evaluation protocol that includes a coarse-grained consistency evaluation, as well as a finer-grained annotation of the responses on various social and emotional criteria. Our study shows that predicting a sequence of expected strategy labels and using this sequence to generate a response yields better results than a direct end-to-end generation scheme. It also highlights the divergences and the limits of current evaluation metrics for generated content. The code for the annotation platform and the annotated data are made publicly available for the evaluation of future models.
\end{abstract}


\section{Introduction}
New, powerful Large Language Models (LLMs) have widely democratised the use of text generation systems, spurring the field of Natural Language Processing toward a new era marked by attempts at reducing the gap between academic progress and day-to-day applications. Such use cases include motivational interviews \cite{galland2023seeing}, customer service \cite{vanelacii} or assistance in psychotherapy sessions \cite{inproceedings}. However, as these models are currently data-driven and generate textual content in a fully end-to-end manner \cite{clavel2022socio}, it is unsure how the social and emotional aspects of the responses formulated by these models, such as informing or sympathising, are planned and regulated. This work aims to join in the effort of building more trustworthy conversational systems. 

\begin{figure*}
\centering
\label{fig:visua}
\includegraphics[width=1\textwidth]{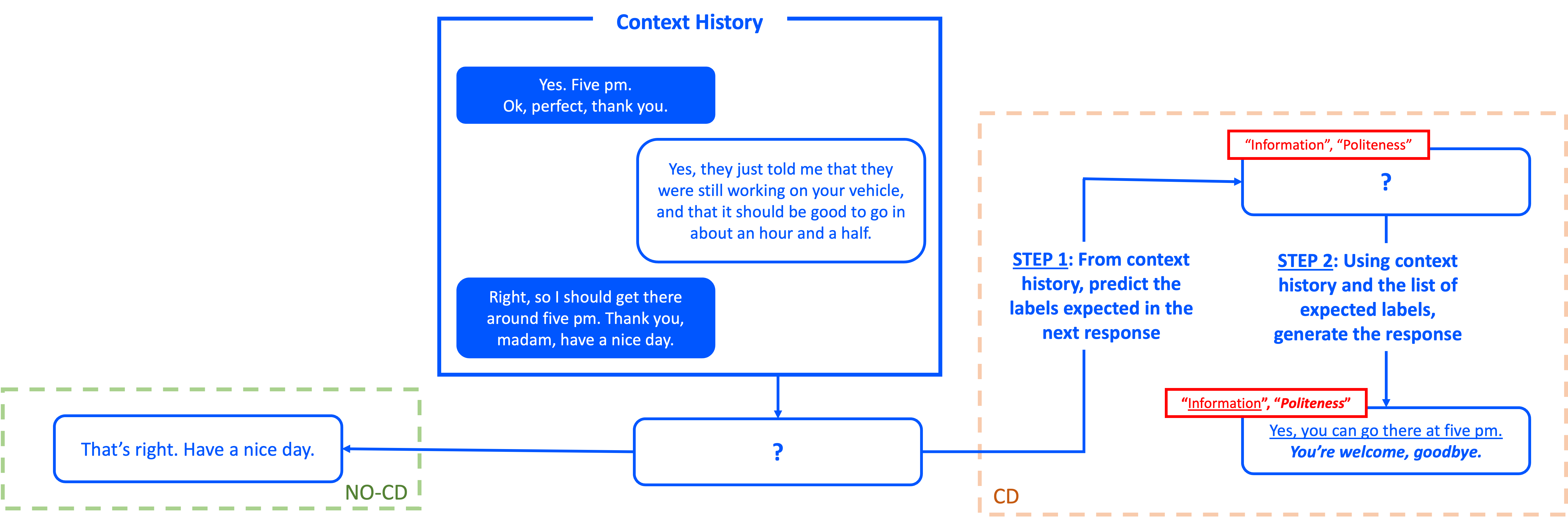}
\caption{Visualisation of the conditional generation approach. NO-CD situation refers to direct response generation, while CD illustrates the conditioning scheme used in this paper.}
\end{figure*}

The contributions of this paper are threefold: \textbf{1)} We propose a response generation system that jointly addresses both the planning and the generation aspects of the process within a neural architecture. As illustrated in Figure \ref{fig:visua}, the process is articulated around two main steps: first, previous turns of the conversation are used to predict a sequence of multiple social and emotional labels. This sequence is then used to condition the selection of the final textual response, by re-ranking a set of generated candidate answers. However, the generation of social and emotional content naturally raises the question of the evaluation. As no automated metrics have yet to properly measure such factors. To provide a dependable analysis of our results, \textbf{2)} we describe our extensive human evaluation protocol that defines multiple criteria that make up the "quality" of an answer. Lastly, \textbf{3)} we share all the code and the annotated data to provide a baseline for future works in the field.\footnote{The code and annotated data are shared in \href{https://github.com/lvanel/response-planning/tree/main}{this repository}.}

\section{Related Work}
\subsection{LLMs for Planning socio-conversational Response Generation}
Response planning is a crucial aspect of building effective and engaging dialogue systems, as it directly impacts the system's ability to maintain natural and contextually coherent interactions. Although end-to-end Large language models (LLMs) have demonstrated impressive skills, particularly in generating fluent text responses, they encounter difficulties with planning tasks. Fully end-to-end approaches such as \cite{kim2023soda} rely on the generation of data controlled by knowledge bases to fine-tune end-to-end models to implicitly integrate socio-emotional strategies. This type of approach gives no visibility or control over the socio-emotional strategy underlying the response, which raises questions of transparency. This is why we have chosen to focus on approaches that provide greater visibility by adding an explicit planning stage. 

Numerous research works have been undertaken for planning socio-emotional strategies either by prompting LLMs or by fine-tuning them. The most advanced prompting approaches rely on Chain-of-thought techniques that are prominent in the literature. Works such as \cite{lee2023investigating} and \cite{li2024enhancing} propose prompting schemes to enhance LLMs' empathetic reasoning skills. The LLMs are thus led to reason in multiple steps on what emotions they should display next to generate the next utterance accordingly. However, such prompt-based approaches rely on a huge amount of data whose content is not always known to decide on the strategy to adopt, which raises problems of explicability (how and why has this strategy been chosen?) and control for task relevance.

Another approach consists of fine-tuning transformer models such as BERT or BART on annotated dialogue data for the planning task. Thus, in \cite{abulimiti2023kind}, various models are fine-tuned to predict the need for a hedge in the next turn, by taking as input a representation of the dialogue history that includes features such as conversation strategies, tutoring strategies or dialogue acts. This "next utterance hedging" prediction is binary (a turn can be either a hedge or a non-hedge turn). \cite{yang2024enhancing} introduces a hybrid approach that combines the task-specific efficiency of smaller empathetic models with the large generative capabilities of LLMs. Thus, a small-scale empathetic model is fine-tuned to predict the most probable emotion (out of the 32 emotions present in the EmpatheticDialogues dataset \cite{rashkin2019towards}) to be generated in the next speaker turn. Then, an LLM is prompted to generate the next utterance conditionally to this emotion.

In this paper, we define an architecture comparable to \cite{yang2024enhancing} that integrates a planning step using fine-tuning approaches followed by a conditional generation step. We propose to improve the planning phase by planning a sequence that combines emotional strategies and conversational strategies to enhance the quality of generated responses. 

\subsection{Socio-Conversational System Evaluation}
In their survey, \cite{Raamkumar_2023} lists the most used metrics for evaluating Empathetic Conversational systems and shows that Perplexity (PPL) is the most popular metric, closely followed by BLEU. Other approaches, such as n-gram-based or sentence-embedding similarity metrics are also commonly used. The survey also insists on the importance of human evaluation to properly evaluate specific user perception-related metrics \cite{li2024enhancing,lee2023investigating}. 

\cite{finch-etal-2023-dont} boasts a high number of labels and seems to provide a comprehensive evaluation of a response’s overall quality. However, their protocol lacks nuance and detail when it comes to what we seek to study which is social and emotional consistency. Evaluating social and emotional content is vast; many aspects can be modelled and defined as evaluation criteria. Studies have focused on the evaluation of \emph{fluency} \cite{li2021knowledge,yang2024enhancing}, \emph{relevance} \cite{yang2024enhancing,thoppilan2022lamda} and \emph{empathy} (defined as emotion appropriateness) \cite{yang2024enhancing,lee-etal-2022-gpt}. We decided to select a set of criteria inspired by these works, which include consistency (derived from the relevance criterion), fluency, and emotion adequacy (derived from the empathy criterion). We also define a new criterion based on social aspects: social adequacy. 

After defining the evaluation criteria comes the question of the evaluation method, particularly how to make the results reusable for comparison with future papers. Different methods exist to evaluate a response. Pair-wise or multiple choice testing \cite{shin2021generating} allows for a strong comparison between available models but makes it hard to compare with future models not considered during the ranking. Alternatively, rating-based systems such as a binary scale or a Likert scale \cite{li2021knowledge,Li_2021} are easier to benchmark, with Likert scales providing a more nuanced evaluation. 

However, for these annotations to be reliable, a good amount of data must be annotated, preferably by more than one human annotator, which amounts to an ever-growing evaluation cost. Evaluation costs also take into account the annotators' cognitive workload during the task. A popular method to decrease such costs is the use of semi-automatic annotation, which can for example entail training a classification model to pre-fill or assist the evaluation \cite{lu2021retrieve,welivita2021large}. 

We use a similar approach in our protocol, but we also choose to divide the evaluation process into steps. First, a coarse relevance filter is used to eliminate the responses that are irrelevant to the context and determine the best three responses among the remaining viable options. Then, only these best responses are rated on finer socio-emotional criteria, reducing evaluation costs while assessing response generation influenced by social and emotional strategies.

\section{Proposed Architecture for the Conditioning by Socio-Emotional Strategies}
\label{section:approach}
The architecture we propose in this paper is composed of two modules, as illustrated on the right of Figure \ref{fig:visua}: a first model is dedicated to predicting the sequence of socio-emotional strategies that the agent is expected to follow in the next speaker turn. Then, in the second module, this sequence is fed to a generative LLM to condition the selection of a final response from a set of generated candidate answers.

\subsection{First Module: Next Strategies Prediction}
Our goal is to develop a planning module to condition and control the next response generation for more socially relevant answers in dialogue. In particular, we are interested in the planning of two specific aspects of conversational strategies we will now refer to as socio-emotional strategies \cite{welivita2021large,icaart23vanel}: \textbf{Emotion-based strategies} (\emph{i.e} expressing happiness or anger) refer to approaches that involve the expression of emotion in response to a user’s emotional state \cite{lin2019caire,emowoz}. \textbf{Dialogue strategies} (\emph{i.e} informing, questioning) are a set of actions and behaviours used to express a conversational intent or goal \cite{Liu2021TowardsES,welivita2021large}. 

We consider the dialogue history of a conversation $C = (c_{i})_{i \in [1, t]}$,  $c_t$ the current speaker turn, and $\mathcal{SE}$ the list of socio-emotional labels. We are interested in predicting the succession of the socio-emotional labels expected to be displayed in the speaker turn  $c_{t+1}$. We thus want to predict the following sequence: $y_{t+1} = (y_{t+1}^{j})_{j \in [1, l_{t+1}]}$ where $y_{t+1}^{j}\in \mathcal{SE}$ and $l_{t+1}$ is the length of the sequence. To determine what model to use, we compared various prompt-based and fine-tuning approaches. We opted to use a fine-tuned BART Base model as it provided the best results on the Daily Dialog dataset (see Appendix A). It predicts on average 1.15 labels per utterance (min: 1 label, max: 3 labels), against the dataset ground truth's 1.20 labels per utterance on average.

\subsection{Second Module: Socio-Emotional Response Generation}
Once obtained, this sequence of labels $y_{t+1}$ is used to condition the generation of the next speaker turn. Two types of methods are investigated here: \textit{i)} A prompt-based approach where LLMs are instructed to generate a response given a 3-turn dialogue history and the expected sequence of socio-emotional strategies; \textit{ii)} a reranking approach such as in \cite{abulimiti2023kind}. For each test sample, a generative model receives the last 3 turns of the dialogue history as input and generates multiple alternative answers (N = 10). To identify the labels present in the generated candidate speaker turns, we train a BERT classifier\footnote{trained for 20 epochs, with a batch size of 32 and a learning rate of 3e-5} on the Daily Dialog dataset.\footnote{On Daily Dialog test set, the scores of the classifier on current utterance multi-label classification are: Jaccard score: 0.59, Precision: 0.70, Recall: 0.76, F1 score: 0.72. The confidence threshold used for the prediction of the current utterance's labels is 0.7, under which the prediction is not considered viable.} Each candidate is fed to this BERT classifier and the resulting list of labels, $l_k$, is compared to the sequence of expected labels, $y_{t+1}$. For each candidate k, we use the Normalised Levenshtein Similarity (NLS) to obtain a similarity score between the socio-emotional labels $_k$ predicted by the BERT classifier and the expected labels $y_{t+1}$ predicted by the first module using the context history. The candidate with the highest similarity score is selected as the final response: $argmax_k NLS(l_k, y_{t+1})$. The conditioning of the response is meant to guarantee that the model generates adequate content that is consistent with both the interaction's context and the social and emotional context of the user. 

\section{Experimental Protocol}
We design an experimental protocol to answer our research question: \emph{Does conditional generation improve the quality of the response?} To that end, we use generative models to compare responses generated without conditioning (\textbf{no-CD}, \textit{i.e.}, considering the first most probable speaker turn outputted by the generative model) and those planned with socio-emotional strategies: \textit{i)} \textbf{CD-pred} using the labels predicted by the first module; \textit{ii)} \textbf{CD-GT} using the same labels as the ones of the ground truth (that are the ones of the human speaker turn in the test set). For each context, the models first generate 10 responses. Then, we follow the method described in Section \ref{section:approach} to rerank the set of candidates and select the one that best matches the expected socio-emotional labels as our final response.

\subsection{Experimental Setting}
\paragraph{Models} 
We compare the following models\footnote{In this work, we used a llama-based model, Beluga, which gave excellent results. But we did not use the later models that came along after we had started the long and thorough human evaluation process. Nor did we use the GPT 3.5+ models because we wanted to promote reproducible research using open-source, freely usable solutions.} (details in Appendix B): \\
\textbf{GPT-2} We fine-tune both \emph{GPT-2 Small} (117M parameters) and \emph{GPT-2 Medium} (345M parameters) \cite{Radford2019LanguageMA}. \\
\textbf{DialoGPT} We fine-tune both \emph{DialoGPT Small} (124M parameters) and \emph{DialoGPT Medium} (355M parameters) \cite{zhang2020dialogpt} to generate an answer given a dialogue context.\\
\textbf{BART} Like in the first experiment, we consider both \emph{BART Base} (140M parameters) and \emph{BART Large} (406M parameters).  \\
\textbf{Beluga} Lastly, we use Beluga (13B parameters) for the prompt-based alternative. We try two approaches: \emph{i) Beluga R (Reranking)}: We instruct Beluga to generate N = 10 responses for each test sample. This is to test Beluga's generation on the reranking approach, comparable to the other models. \emph{ii) Beluga PB (Prompt-Based)}: We directly instruct Beluga to generate a response to the 3-turn context using a certain tone conditioned by the expected labels. For task CD-GT, the expected labels are the ground truth labels from the dataset, and for task CD, the labels are predicted by the BART model.

We fix a context window of 3: we select 3 speaker turns to predict the next labels sequence and the same 3 turns to generate the next. We slide the window over each conversation to obtain all the different sets of 3 turns for every conversation.

\paragraph{Data}
The models were trained and tested on the DailyDialog dataset \cite{li2017dailydialog}. DailyDialog consists of scripted dialogues of typical conversations designed to help people learn English. We thus expect the strategies and emotions, annotated by humans, to be adequate and respect the commonly accepted social norms. This dataset is in English and thus mainly represents the social customs of the English-speaking world. We chose this dataset as it is one of the only publicly available conversational resources annotated with both emotions and dialogue acts.

\subsection{Automated Evaluation}
To evaluate the quality of the generated responses, we use various metrics implemented in the HuggingFace \emph{evaluate} library. We use string-based metrics (\emph{Sacrebleu} \cite{post-2018-call}, \emph{Rouge} \cite{lin-2004-rouge} and \emph{chrf} \cite{popovic-2015-chrf}), as well as embedding-based metrics (\emph{BERTscore} \cite{bert-score} (between the generated candidate and the reference)) to measure the quality of the generated content. Both approaches are based on a comparison of the generated content to the dataset reference. We also look at reference-free metrics: \textit{i)} the \emph{BERTScore} to measure the distance between the generated candidate and the context history (BERTscore context), \textit{ii)} \emph{Perplexity} (PPL) \cite{jelinek1977perplexity} that measures how well a language model predicts a text sample.

\subsection{Human Evaluation}
While these automated metrics are convenient and easily accessible, most of them are dependent on the reference which makes them obsolete when it comes to evaluating tasks such as response generation: to the same context, many responses can be appropriate, even if they are very different from the ground truth.  

To obtain dataset-independent results that reflect this fact, we perform a human evaluation on a randomly selected sample of 300 contexts extracted from the test set. After running each \{model, conditioning\} combination over our test dataset, a list of 23 generated responses per context is obtained, to which the human reference found in the dataset is added. Since the CD-GT and CD-pred conditioning methods rely on a reranking approach based on the same pool of 10 generated candidates, they can often select the same candidate. For each context, the duplicates are thus removed. 

The annotation process is divided into three steps, to reduce the workload for the annotators. First, the responses associated with the same context are divided into those that are consistent and those that are not. Second, the best responses among the consistent ones are selected by the annotators. Third, once only the "best" responses remain, they will then be annotated with more precise criteria. 

\paragraph{Step 1: Filtering}
For this task, every unique response is displayed, and the annotators must filter out the responses that are not relevant to the context. At the end of this phase, the set of answers is divided between those who have been eliminated (were not consistent with the context) and those who have been validated (judged as viable and consistent). Thus, for each context, the human annotators are asked to evaluate all responses on the two following criteria: \emph{Consistency} evaluates whether a model's response makes sense with the context (for example, if it does not contradict the context or is off-topic) and \emph{Specificity} measures how specific the response is to the context (for example, if the context is "I love tea." the answer "Same" could be a response to many other conversations, but "I do too, especially black tea." is more specific.)

\paragraph{Step 2: Top-3}
Once the filtering phase is over, the annotator must pick the best three answers from the pool of relevant responses. This choice is based on consistency to the context as well as how specific the response is. There is no notion of order among this top-3, it just aims to identify the three best available options among the set of generated answers.

\paragraph{Step 3: Socio-Emotional Annotation}
The final step of the evaluation protocol was designed to annotate the best responses to all contexts with finer-grained socio-emotional criteria. To that end, for each context we consider the union of the top-3 selected by both annotators which allows us to work with a smaller pool of responses and only annotate the answers that were approved and picked by the human experts. We consider the union rather than the intersection of the top-3 to keep the most answers that the annotators approved and obtain multiple ratings for one context.

First, the response is pre-annotated by a BERT classifier trained to predict a label among the four dialogue acts found in the Daily Dialog dataset. The annotator corrects this prediction and then rates the response on different questions. These questions are sorted into three axes: \emph{logical consistency} (usefulness of the answer, fluency, style consistency), \emph{emotional consistency} (is the emotional tone of the response adequate?) and \emph{social consistency} (adequacy of the dialogue strategies, role consistency). The complete description of the questions and the rating scale are described in Appendix D. This allows us to obtain multiple annotated responses for the same context. We also use these annotations to investigate the efficiency of the conditioning approach. In this part, for reasons related to time and resource costs, the three annotators annotated the same set of 59 contexts, which represents about 250 individual candidate responses.

The \emph{gradio} library \cite{abid2019gradio} was used to develop an annotation platform specific to the task (the code to the platform is available on the GitHub repository linked in the introduction). As human evaluation yields more qualitative results than using mass crowd-sourcing platforms, we decided to request the help of human experts fluent in English. The three annotators are women who have obtained a master's in linguistics or NLP and who work as conversational data analysts and annotators. For Steps 1 and 2, each context and its set of responses are evaluated by two annotators, to compute an inter-annotator agreement. The annotators do not know which response corresponds to what model input and do not know which response is the human reference extracted from the dataset. The responses are shuffled randomly for every sample, to avoid the creation of any pattern or bias. For step 3, all three annotators evaluate the same subset of responses. For all steps, a sample of 10 contexts was first evaluated, to allow the three annotators to discover the tasks and become familiar with the evaluation criteria. When this first test run was achieved, they evaluated the rest of the sample. 

To compute an inter-annotator agreement meant to measure the overlap of responses filtered as relevant, we look at Krippendorff's Alpha, as it provides an agreement measure that supports multi-label (we want to compare the list of responses, in other words, the list of {models, conditioning} combinations "saved" by both annotators). The alpha is computed for all three annotator pairs and yields an average inter-annotator agreement of 0.51, which is very satisfactory\footnote{For reference, the SODA dataset \cite{kim2023soda} obtained a Krippendorf’s alpha of 0.25 between 74 annotators.}. We also look at the Jaccard distance of the two lists and obtain a score per annotator pair, that averages a similarity of around 0.97.

\paragraph{Evaluation Scores}
After gathering the annotations on each step of the evaluation process, we compute various metrics to analyse the results. We call N the number of contexts annotated, k the number of annotators, and m a model.

\emph{(Step 1)} For each annotator, we compute $filter\_i(m)$ that refers to the number of times the responses of model m were filtered as "consistent" by an annotator i. We look at the average score: $filter(m) = \frac{1}{k} * \sum_{i=1}^{k} \frac{filter\_i(m)}{N} $.  

\emph{(Step 2)} We also look at how often its response is chosen as part of a top-3, with $top3\_i(m)$ the number of times the model's responses have been selected in the top-3 by an annotator i, and compute the average score: $top3(m) = \frac{1}{k} * \frac{\sum_{i=1}^{k} top3\_i(m)}{N} $. 
   
(For the first two steps, N = 300 and k =3.) These percentages are shown in the first two columns of Table \ref{tab:results_conditioning}. 

\emph{(Step 3)} Using the socio-emotional evaluations obtained in the previous step, a response can be attributed to a score on all three axes: a logical consistency score, an emotional consistency score and a social consistency score. To compute these axes scores, we take the rating of each question of the category, normalise them, and calculate the mean score. However, we are interested in a single, more global consistency score that evaluates the quality of the response to the logical, emotional and social context of the interaction. Thus, for each model, we leverage the mean of these specific consistency scores and weigh this score by the number of times the model's responses were chosen. We define the \emph{socemo} index such as:
\[socemo(m) = \frac{1}{N} * \frac{\sum_{i=1}^{k} logi\_i(m) + emo\_i(m) + soc\_i(m)}{k}\] (In our case, N = 59 for this step. $logi\_i(m)$, $emo\_i(m)$ and $soc\_i(m)$ the logical consistency score, an emotional consistency score and a social consistency score annotated by the annotator $i$ for the model m.)

\section{Results \& Discussion}
\begin{table*}[ht]
    \centering
    \begin{small}
    \label{tab:dialogue}
    \begin{tabular} {p{36mm}|p{5mm}|p{5mm}|p{5mm}||p{6mm}|p{8mm}|p{7mm}|p{7mm}|p{12mm}|p{6mm}|}
        \hline
        \textbf{Model} & \textbf{filter} & \textbf{top3} & \textbf{soc emo} & \textbf{Sacre bleu} & \textbf{Rouge} & \textbf{Bert score} & \textbf{CHRF} & \textbf{Bertscore context} & \textbf{PPL} \\ 
        \hline 
        \emph{GPT-2 \textsubscript{b} NO-CD} 	    & 33&9&13  & 76    & 12    & 84      & 13.6 & 85 & 845\\
        \emph{GPT-2 \textsubscript{b} CD-pred} \textsubscript{(R)}     	    & 37&9&14 & 100     & 13    & 85      & 13.4 & 85 & 204  \\
        \emph{GPT-2 \textsubscript{b} CD-GT} \textsubscript{(R)}     	    & 37&9&14 & 100     & 12    & 85      & 13.4 & 85 & 204  \\
        \hline   
        \emph{GPT-2 \textsubscript{M} NO-CD} 	    &\textbf{53} &21&28 & \textbf{192}    & 14    & \textbf{87}      & 14.1 & 85 & 86 \\
        \emph{GPT-2 \textsubscript{M} CD-pred} \textsubscript{(R)}     	& \textbf{5}&19& 30 & 184    & 14    & \textbf{87}      & 15.8 & 85 & 85 \\
        \emph{GPT-2 \textsubscript{M} CD-GT} \textsubscript{(R)}    	    & \textbf{5}&19& 30 & 184    & 14    & \textbf{87}      & 15.8 & 85 & 85 \\
        \hline  \hline   
        \emph{DialoGPT \textsubscript{b} NO-CD}     & 37&11&18 & 120    & 13    & 85      & 12.7 & 85 & 84 \\
        \emph{DialoGPT \textsubscript{b} CD-pred} \textsubscript{(R)}     	& 4&13 &19 & 122    &	13    & 85      & 15.4 & 85 & 72 \\
        \emph{DialoGPT \textsubscript{b} CD-GT} \textsubscript{(R)}    	& 4&13 &19 & 122    &	13    & 85      & 15.4 & 85 & 72\\
        \hline
        \emph{DialoGPT \textsubscript{M} NO-CD}    & \textbf{53}&16&15 & 151    & 14    & \textbf{87}      & 13.0 & 85 & 81\\
        \emph{DialoGPT \textsubscript{M} CD-pred} \textsubscript{(R)}      & \textbf{52}&16&20 & 151    &	14    & 85      & 15.9 & 85 & 71  \\        
        \emph{DialoGPT \textsubscript{M} CD-GT} \textsubscript{(R)}       & \textbf{52}&16&20 & 151    &	14    & 85      & 15.9 & 85 & 71  \\
        \hline         \hline
        \emph{BART \textsubscript{b} NO-CD} 	    & 32&7&13 & 113    & 13    & 86      & 10 & 86 & 62 \\
        \emph{BART \textsubscript{b} CD-pred} \textsubscript{(R)}	& 32&7&14   & 146    &	13    & \textbf{87}      & 15.9 & 86 & 65  \\
        \emph{BART \textsubscript{b} CD-GT} \textsubscript{(R)}	    & 32&7&14 & 146    &	13    & \textbf{87}      & 15.9 & 86 & 65  \\
        \hline
        \emph{BART \textsubscript{L} NO-CD} 	& 42&9&19   & 129    & \textbf{17}     & \textbf{87}     & 10.2 & 86 & 62\\
        \emph{BART \textsubscript{L} CD-pred} \textsubscript{(R)}	& 45&12&19    & 151    &	14     & \textbf{87}     & 16.0 & 86 & 58  \\
        \emph{BART \textsubscript{L} CD-GT} \textsubscript{(R)}	& 45&12&19    & 151    &	14     & \textbf{87}     & 16.0 & 86 & 58 \\
        \hline         \hline
        \emph{Beluga NO-CD} 	    & 42&25&39  & 96    & 12   & 85       & 15.8 & 85 & 5624\\
        \hline
        \emph{Beluga R CD-pred} \textsubscript{(R)}	& 3&1& \emph{NA}    & 84    & 10   & 84       & 11.7 & \textbf{90} & 7497\\
        \emph{Beluga R CD-GT} \textsubscript{(R)}	& 3&1& \emph{NA}    & 84    & 10   & 84       & 11.7 & \textbf{90} & 7497 \\
        \hline
        \emph{Beluga PB CD-pred} \textsubscript{(Prompt)}    	& \textbf{51}&\textbf{36} &44    & 89    & 13    & 86      & \textbf{17.9} & 87 & 69\\
        \emph{Beluga PB CD-GT} \textsubscript{(Prompt)}    	    & 45&30 &\textbf{51} & 87    & 13    & 86      & 17.8 & 87 &362 \\
        \hline         \hline
        \textbf{Daily Dialog Reference} 	    & \textbf{87} & \textbf{61} & \textbf{69} &    &    &      & & 85& 132  \\
        \hline
    \end{tabular}
    \caption{Comparative results of the experiments on conditioning response generation using multi-label sequences modelling social and emotional behaviours. Results are given in \%. \textsubscript{b} denotes the Base or Small model, \textsubscript{M} the Medium model and \textsubscript{L} the Large model. \\ \textsubscript{(R)} means the approach used is reranking, while \textsubscript{(Prompt)} refers to prompt-based generation.}
    \label{tab:results_conditioning}
    \end{small}
\end{table*}

Table \ref{tab:results_conditioning} summarises the results discussed in this section. We observe that conditioning yields slightly better results on both automated and human evaluation metrics\footnote{After we had carried out the human evaluation, it was brought to our attention that the official \emph{huggingface} split of the DailyDialog dataset displays duplicates in the test and training set \cite{wen2022empirical}. In Appendix F, we present the results obtained across the automated metrics on all the models trained on a different split of Daily Dialog that does not feature duplicates (the one provided in the original paper). The new results are consistent with those presented in Table \ref{tab:results_conditioning}. They show similar trends between models' scores and that generally CD-pred $=$ CD-GT $>$ NO-CD.}. Then, we see that conditioning on predicted labels does not seem to induce a significant decrease in results compared to the "ideal" ground-truth conditioning. 

\paragraph{Results of the evaluation of the consistency criteria (Evaluation Steps 1-2)}
Out of the 24 available responses, there is an average of 19 considered responses once we have removed the duplicate answers to the same context. 

\emph{STEP 1:} Human evaluation eliminated on average 10 candidates per context, to retain 9. The human reference is consistently better and is deemed as "relevant" 87\% of the time. When we look at the generated responses, we notice that only GPT-2 Medium and DialoGPT Medium, as well as BELUGA PB CD-pred, are saved more than 50\% of the time. 

\emph{STEP 2:} For the top 3, as two annotators judged each set of responses, the overlap shows that the size of the union of the selected top-3 responses is 4, while the average intersection size of the two top-3 is 1.6. The human reference is chosen as part of the top-3 best responses 61\% of the time. CD models tend to do better, with BELUGA CD-pred significantly outperforming the other generative approaches. Some models obtain very low results on this task, namely the Beluga R models, but also the Base / Small models. This second step allows us to mark the gap between the better models (Beluga PB and NO-CD, GPT-2 Medium, DialoGPT Medium) and the rest, highlighting the difference in quality that might not have been as obvious after the first consistency filter in Step 1.

\paragraph{Results of the socio-emotional criteria evaluation (Evaluation Step 3)}
The results of the annotation of fine-grained socio-emotional criteria seem to show that both CD and NO-CD responses, once filtered by consistency, tend to be of equally good quality across all three axes: logical, emotional and social. It is important to keep in mind that for this step, only 59 contexts were annotated out of the 300 considered in the previous steps (around 250 individual responses), so the sample is quite smaller than for the previous task. Beluga R models are not represented as they were very seldom selected in the annotators' top-3.

As specified previously, the $socemo$ score combines both the logical, emotional and social consistency ratings, as well as the frequency with which the model was selected as one of the top-3 best responses to a context amidst the 24 available responses. Overall, Beluga PB CD-GT is the model, apart from the dataset reference, that is the most represented in our sample, followed by Beluga PB CD-pred. The $socemo$ score shows a consistent increase when it comes to CD models compared to their non-conditioned alternative. This mostly comes from the fact that CD models were preferred in the response selection phase (Steps 1-2). Unweighted logical, emotional and social scores, show that once the responses have made it to the top-3, they all present generally good ratings. However, these scores are the result of an evaluation carried out on unbalanced samples where all models were not equally as represented, which is why the $socemo$ score is more reliable. Other than the Beluga models which significantly outperform the others, the larger models seem to be yielding better results, with GPT-2 Medium scoring fairly high.

We also notice that CD-pred results are extremely similar to CD-GT. Both tasks use a reranking approach on the same 10 generated sentences, the only difference being the set of `expected labels'. This shows that even when using a generator to output the sequence of labels expected for the next utterance, the error margin of the generator does not impact the candidate selection results. CD-pred models even tend to have better mean NLS compared to the CD-GT models. This means that on average, there is a higher similarity between the labels generated by BART and the labels predicted by BERT Current for the final candidate. However, when we look at the responses of every model to one test sample, it shows that CD-GT and CD-pred models often select the same candidate, which explains the similar results on all the other metrics. Beluga is the only exception to this rule, as the prompt-based approach means that CD-GT and CD-pred do not select from the same pool of responses. Beluga CD-GT only outperforms CD-pred by 1\% on the total score.

\paragraph{Human metrics against automated metrics}
In some cases, the automated metrics seem to echo some of the results observed by the human evaluation. For example, GPT-2 Medium's performance surprisingly surpass those of DialoGPT on automated metrics. Trained with similar hyper-parameters, DialoGPT is supposed to be better suited to dialogue generation but presents here slightly worse results than GPT-2. While BART Base does better than both DialoGPT Small and GPT-2 Small, DialoGPT Large's performance are on par with BART Large's. Both human evaluation and automated metrics show a clear increase in performance in bigger models. Even though, for computation reasons, we could not train DialoGPT Large or GPT-2 Large, the Medium-sized models yield better results than their Base counterparts. Both approaches also seem to agree that CD-GT and CD-pred models seem to display equivalent performance. For the prompt-based model, we compare the reranking approach with the direct conditioning via instruction. Beluga PB outperforms Beluga F\& R on both CD tasks. Some of the outputs for the Beluga NO-CD and Beluga R models were empty or unparsable, while this issue was not observed with the PB model. 

However, there are also many aspects where human evaluation and automated metrics present divergences. The BERTscore$\_context$ computed between context and response was included to give a measure of similarity or closeness between the two, and hypothesising that it could be equivalent to a measure of logical consistency. When we contrast the results obtained by this score with the results of the filtering and top-3 steps of the human evaluation, we see a disconnect. Where the BERTscore$\_context$ shows the best results with the Beluga R models, which are unarguably the worst-performing models according to human input. This is because the responses generated by these models tend to repeat verbatim parts of the context history, hence the high similarity score.

When looking at the set of automated metrics, the highest results seem to indicate that the best-performing models are GPT-2 Medium or BART Large, but the human evaluation seems to prefer Beluga PB, GPT-2 Medium and DialoGPT Medium models. It is particularly interesting to see how the automated metrics fail to capture Beluga PB models' efficiency compared to the other systems. \cite{nimah2023nlg} draws parallels between Perplexity (PPL) and fluency or how natural a response sounds. While the two are not perfectly equivalent, they both aim, in a way, to evaluate the quality of the construction of the sentence according to a language model. We extracted the fluency score from our human evaluation and weighted it similarly to the $socemo$ score to compare it to the PPL scores (see Appendix E). PPL seems to indicate that the human reference scores a higher (the lower the score, the better) score than most non-Beluga models, except for GPT-2 Small NO-CD. All three GPT-2 Small scores are surprisingly high, but not as high as all the Beluga models, except Beluga PB CD-pred, which presents some of the best perplexities out of the considered models. The PPL results for GPT-2 Small NO-CD and the Beluga models are higher because they're the only models that have generated non-parsable or NaN answers. In general CD models seem to be doing better than NO-CD, a trend that corroborates the human evaluation. However, when it comes to the evaluation of the quality of the models themselves, PPL seems to contradict the human results. The model that seems to be doing the best according to PPL is BART Large, Beluga NO-CD's result is extremely high, and GPT-2's performance is below those of both BART and DialoGPT.

\paragraph{Limitations}
As this article is one of the first to explore the role of socio-emotional conditioning in LLMs, we chose to start with a basic prompt to compare to the other approaches we considered. Using more prompt-based solutions and comparing them to our current benchmark is a longer-term objective. Besides, one of the goals of this paper is to prove that conditioning improves the social and emotional quality of a generated response across various types of approaches (traditional generative and prompt-based systems). While adding newer models might have improved the general results, we do not think their absence disproves our findings. This paper focuses on reproducible results and provides a first baseline with a wide range of models on this novel task, and we would encourage future studies to compare themselves to this benchmark.

\section{Conclusion}
This paper is the first to tackle the task of jointly predicting explicit dialogue and emotion-based strategies to condition response generation using LLMs. This novel approach requires the release of new resources, which is why we propose a dual contribution: First, we propose an architecture to condition the response generation by a set of dialogue and emotion-based strategies. Then, to properly evaluate our approach, we describe a new human evaluation protocol for socio-emotional response generation and introduce a novel criterion for social adequacy. This protocol aims to reduce the annotation costs without sacrificing the evaluation's depth and precision and is validated by a satisfactory inter-annotator agreement. The details are presented in Appendix D and both the code for the evaluation interface as well as the data (samples of the Daily Dialog dataset) annotated by our team are shared for comparison of future models.

The evaluation leads to two main results: \textit{i)} \textbf{Conditioning improves the quality of the generated response,} both on the general consistency of the answer, as well as the finer-grained social and emotional criteria. Conditioning on dataset labels is often equivalent to conditioning on predicted labels, which means that even if the intermediate step does perfectly predict the sequence of labels, it is close enough to obtain results similar to the ideal dataset-assisted scenario; \textit{ii) }\textbf{Contrasting automated metrics and the human results} show that while automated metrics manage to pick up some general trends of the quality evaluation, they are still unable to capture important information, especially when it comes to social behaviours. Current automated metrics do not suffice to properly evaluate the quality of a response.

Our future work includes exploring new LLM-based conditional generation approaches and comparing them to the baseline established in this paper, to develop a dialogue system able to generate responses that are both context-relevant as well as socially and emotionally consistent. We also mean to investigate the influence of planning emotions and dialogue strategies individually to explore their individual contribution to the socio-emotional quality of the response.

\section*{Acknowledgment}
This work was partially funded by the ANR-23-CE23-0033-01 SINNet project. We thank our team of conversation analysts at Zaion for their diligence and hard work in carrying out the evaluation of the responses, allowing us to provide insight and reliable results to this study.

\section*{Ethical Impact Statement}
This study features an evaluation carried out by a team of human linguists, and it is important to note that annotation includes biases. They can be related to the personal and cultural experiences of the annotators, which may influence their perception of emotions and interactions. Thus, the data we provide in this study may contain some biases on how emotion is perceived and labelled, but communication and reference materials were shared between all annotators to curb these differences as much as possible.

On another note, modular architectures allow for a more explicit selection of dialogue policies, which is not the case in end-to-end approaches. The current NLP trends seem to favour end-to-end approaches, especially as large LLMs have proved their proficiency, but it often comes at the price of transparency. This research seeks to find a middle ground between the more rigid architecture of modular systems with the computation power or larger end-to-end solutions while trying not to compromise transparency.

We aim to develop a system that can accurately generate a response that matches the social and emotional tone of the user's utterances as well as the context, but we want to do so in the most transparent way possible, with a model able to justify its output with understandable arguments. It is also important to note that in the realm of conversational AI agents equipped with social and emotional capabilities, a noteworthy emerging risk lies in their potential to sway consumers towards making purchases or believing misinformation. 

To our knowledge, this paper is the first to propose an approach that combines explicit planning with LLMs. It also involves jointly predicting emotion-based strategies as well as dialogue strategies, which we haven’t seen being done in the literature. To coordinate these two novel concepts into a single architecture, we first opted for a system with simple, explicit modules to supervise the two steps of the process (planning, and then generating), before we can move on to more complex alternatives. We are also aware that it is crucial to study cultural differences when it comes to social interactions. However, the lack of resources that include both emotion and dialogue acts annotations as well as culturally rich dialogues currently does not allow us to provide a reliable generalisability on this aspect.

While we are working on this subject to contribute to the scientific community and to improve the quality of the service that agents offer by being more tuned to the users’ emotional and social situations, we are aware that it could be used in defective ways. We believe that communicating and informing the users of such systems is crucial to developing their awareness of such potential risks, as well as protecting them for their future interactions with AI systems. 

\bibliographystyle{IEEEtran}
\begin{small}
    \bibliography{custom}
\end{small}

\section*{Appendix A: Choice of the Next Label Prediction Model}
\label{sec:nextlabelspred}

In this first step, we aim to evaluate the performance of various models on the task of predicting a sequence of labels that models the social and emotional behaviours that are expected to be displayed in a generated response to a conversational context. In other words, we want to test the first step of our approach and determine the most suitable model to use as the planning module. 

\paragraph{Data Preprocessing}
We work with the Daily Dialog dataset. For each speaker turn, we consider 3 dialogue turns as the "context" and pair them with the label(s) of the following utterance to constitute a training sample. The model thus learns how to predict the labels of the next speaker turn. Our resulting train/validation/test splits are made up of 76052 / 7070 / 6740 samples.

\renewcommand{\arraystretch}{1.4}
\begin{tabular} {|p{40mm}|p{25mm}|}
    \hline
    \textbf{Utterance} & \textbf{Labels} \\
    \hline
    You surely know a lot about Chinese tea. & inform \\ \hline
    Sure, I like drinking tea at teahouses. & inform, happiness \\ \hline
    Oh, so do I. & inform \\ \hline
    Why don't we go for one now? & directive \\ \hline
    Great. We can chat while enjoying a cup there. & commissive, happiness \\ \hline
\end{tabular}
\label{tab:results_labels}

For such a conversation, we can extract a few training samples, for example: \\
\textbf{Context}: 'Sure, I like drinking tea at teahouses. | Oh, so do I. | Why don't we go for one now ?' \\
\textbf{Labels}: 'commissive, happiness'

\paragraph{Models}
All the models we present were trained using a single GPU (NVIDIA RTX 8000, 48GB memory), with the hyper-parameters described in the Appendix \ref{sec:hyperparams}. We describe the models used in Experiment 1 below: 
\paragraph{BERT - Multilabel Classification} \emph{BERT Base} (110M parameters) and \emph{BERT Large} (340M parameters) \cite{devlin2019bert} are trained on a multi-label classification task. We set the confidence threshold at 0.7 for BERT Base and 0.5 for BERT Large.
\paragraph{BART - Sequence Generation} \emph{BART Base} (140M parameters)  and \emph{BART Large} (406M parameters) \cite{lewis2019bart} are fine-tuned on the task of generating the next labels sequence.
\paragraph{Beluga - Prompt-Based Generation} We use Beluga (13B parameters), a Llama2 model \cite{touvron2023llama} fine-tuned on an Orca style dataset, to generate the sequence of the next labels using few-shots prompt-based generation. \emph{Beluga} was prompted to generate the sequence of labels associated with the following speaker turn, given a dialogue utterance. The prompt used is:
\emph{
Predict the sequence of labels associated with the utterance that follows the given dialogue.\\
We consider the following labels: `inform', `question', `directive', `commissive', `neutral', `anger', `disgust', `fear', `happiness', `sadness' and `surprise'. The answer must be one or a sequence of multiple labels from this list.\\
\\
Here are a few examples,\\
Dialogue: Good morning, sir. Is there a bank near here ?\\
Labels: `inform'.\\
Dialogue: Is it far ?\\
Labels:`inform'\\
Dialogue: No, It's only about five minutes walk.\\
Labels: `inform', `happiness'.\\
\\
What labels are associated with the utterance following this dialogue: \\
Dialogue: + [current utterance]}

\textbf{Random Selector - Baseline} We add a random selector that will, for each utterance, select random labels out of the list of available labels. This model is meant to serve as a comparison with the other two models. We randomly select k labels out of the list, k chosen randomly between 1 and 2, following the length distribution observed in the dataset.

\paragraph{Metrics}
As a sub-task of the response generation process, label sequence prediction is a one-to-many problem: many sequences can match a same context. However, efficiently evaluating the relevance of a sequence of labels to a context remains a challenging task due to the lack of suitable metrics. Thus, to evaluate this experiment, we must rely on comparing the pairs of sequences: the generated or predicted sequence, and the expected sequence. 

To evaluate the results, we rely on metrics implemented in the \emph{scikit-learn} library, such as the \emph{Jaccard Score}, used to compare two sets of labels to evaluate the similarity between the predicted set and the expected set, or the multi-label implementation of \emph{F1 score, Precision, Recall} that allows us to measure the performance of the models against the expected sets of labels. For the sequence generation task (BART and Beluga), we also measure the \emph{Normalised Levenshtein Similarity (NLS)} Levenshtein Distance ($LD$), a lexical similarity measure which identifies the distance between one pair of strings. It represents the smallest number of base edit operations, namely insertion, deletion or substitution, required to transform the source sequence $S$ into the target sequence $T$. Levenshtein Similarity ($LS$) is computed as $LS = 1 - LD$, and it is normalised as $NLS = (1 - LD) / max(len(T), len(S))$. Normalised Levenshtein Similarity is implemented in the \emph{textdistance} library. Lastly, we look at the mean length of the generated/predicted sequences, \emph{Mean $l_{i}$}, to contrast it with the dataset's average of 1.20 labels per utterance.

These metrics are efficient in comparing the gap between what is predicted and what labels were used in the real conversation, but it is important to keep in mind that when it comes to dialogue there is not one single good answer. There are many different ways to participate in a conversation, and there is no guarantee that a different agent would have used the same strategies.
\begin{table*}[ht]
    \centering
    \begin{tabular} {|p{15mm}|p{12mm}|p{12mm}|p{12mm}|p{12mm}|p{12mm}|p{12mm}|}
        \hline
        \textbf{Model} & \textbf{Jaccard Score} &\textbf{Preci-sion} & \textbf{Recall} & \textbf{F1 Score} & \textbf{NLS} & \textbf{mean $l_{i}$}\\
        \hline \hline
        \emph{BERT\textsubscript{b}} 	    & 0.34  &   0.43  & 0.62  &	0.49 & \emph{NA}   & 0.58\\ 
        \hline
        \emph{BERT\textsubscript{L}} 	& 0.38  &	1.00  & 0.38  &	0.55 & \emph{NA}   & 1.00\\
        \hline \hline
        \emph{BART\textsubscript{b}} 	    & 0.38  &	0.56  & 0.53  &	0.54 & 0.54   & 1.15\\
        \hline
        \emph{BART\textsubscript{L}} 	& 0.38  &	0.54  & 0.54  & 0.54 & 0.53   & 1.22\\ 
        \hline \hline
        \emph{Beluga}   & 0.020 &	0.04  & 0.05  & 0.04 & 0.099  & 2.72 \\
        \hline \hline
        \emph{Random} 	    & 0.035 &	0.11  & 0.10  &	0.07 & 0.12 & \textbf{1.20} \\ 
        \hline 
    \end{tabular}
    \caption{Comparative results of the experiments on conditioning the generation of a multi-label sequence of social and emotional behaviours. \textsubscript{b} denotes a Base model, and \textsubscript{L} indicates a Large model.}
    \label{tab:results_labels}
\end{table*}

\begin{table*}[ht]
  \centering
  \begin{tabular} {|p{25mm}|p{15mm}|p{15mm}|p{15mm}|p{15mm}|}
    \hline
    \textbf{Model} & \textbf{Trained Epochs} & \textbf{Learning rate} & \textbf{Batch-size} \\
    \hline
    \textbf{BERT}            &    &     &  \\
    \hspace{3mm}\emph{Bert Base} 	  & 10   &	3e-5  & 32 \\
    \hspace{3mm}\emph{Bert Large} 	  & 10   &	3e-5  & 32 \\
    \hspace{3mm}\emph{Bert Current} 	& 20   &	3e-5  & 32 \\
    \hline
    \textbf{BART}            &    &     &  \\
    \hspace{3mm}\emph{BART Base} 	  & 10   &	3e-5  & 32 \\
    \hspace{3mm}\emph{BART Large} 	  & 10   &	3e-5  & 32 \\
    \hline
  \end{tabular}
  \caption{Hyper-Parameters for training.}
  \label{tab:hyperparams1}
\end{table*}

As a classifier, BERT operates without inherent awareness of sequence order; it processes input as unordered lists rather than predictive sequences. While we did expect a lower performance, we believed it was interesting to compare a "safer" method, such as classification, that is forced to predict real labels, to generative methods that can be prone to hallucinating. However, when we look into the predictions outputted by the BERT models, we see that it only predicts the main class and does not manage to provide diverse outputs. 

As for the generative approaches, the amount of data required to confidently fine-tune a generation model such as BART is quite demanding, especially on a task such as next utterance labels generation. Prompt-based approaches such as Beluga, through the use of few-shot prompting, offer a more data-efficient approach. We are interested to see if the capabilities of a large, Llama2-like model, can bridge the performance gap with a data-driven model such as BART. BART yields results more interesting and diverse than BERT's. While BART Base displays the best performance, the performance of BART Large does not parallel its larger scale, presenting comparatively inferior results.
As prompt-based models have been rising with the success of ChatGPT, new possibilities have become accessible. However, when it comes to sensible or confidential data, it is hard to use such online services. We picked the Beluga model because it is an open-source alternative, fine-tuned from Llama 2, with good overall performances in English. However, we tried different prompts for Beluga but none were conclusive. The results are far below the previous methods, and even comparable to the low Random model baseline. One of the biggest issues with the prompt-based approach is that many results were not parsable, outputting 'None' as a label sequence even when explicitly told not to do so.

\section*{Appendix B: Details on the models used for Conditional Generation}
\label{sec:conditionalgen}
Here are the main hyper-parameters used to train each model presented in this paper. Each model was trained using a single GPU (NVIDIA RTX 8000, 48GB memory).
\begin{table*}[ht]
  \centering
  \begin{tabular} {|p{25mm}|p{25mm}|p{25mm}|p{25mm}|p{25mm}|}
    \hline
    \textbf{Model} & \textbf{Epochs} & \textbf{Learning rate} & \textbf{Batch-size} \\
    \hline
    \textbf{GPT-2}             &     &     &  \\
    \hspace{3mm}\emph{GPT-2 Small} 	    & 12   & 2e-5    & 32 \\
    \hspace{3mm}\emph{GPT-2 Medium} 	  & 5    & 2e-5	  & 16\\
    \hline  

    \textbf{DialoGPT}            &     &     &  \\
    \hspace{3mm}\emph{DialoGPT Small} 	  & 8    & 2e-5    & 32 \\
    \hspace{3mm}\emph{DialoGPT Medium} 	  & 4    & 2e-5    & 16 \\
    \hline
    
    \textbf{BART}              &     & 2e-5    &  \\
    \hspace{3mm}\emph{BART Base} 	    & 10   &	2e-5    & 32 \\
    \hspace{3mm}\emph{BART Large} 	    & 6    & 2e-5    & 16 \\
    \hline
  \end{tabular}
  \caption{Hyper-Parameters for training.}
  \label{tab:hyperparams2}
\end{table*}

\subsection{Beluga: Prompts for Conditional Response Generation}
Multiple prompts were tested to optimise the results and here are the final instructions used to train Beluga for the two experiments. 
Here, $N$ is the number of sequences to be generated. In this paper, we used N = 10. Element refers to the dialogue history considered, we use a window of 3 utterances of context. We set the dialogue history in the format: SPEAKER A: utt1 SPEAKER B: utt2 SPEAKER A: utt3.\\

\paragraph{Beluga R}

\textbf{For the generation of a single response, `NO-CD' task, the prompt used is:}
Generate the response following the given context. For example:\\
A: Do you like some soup? \\
B: Yes, but I don't know what soup you have \\
A: We have beef soup and tomato soup \\
Response: Good. I prefer beef soup .\\
\\
A: Can I take your order now, Madam? \\
B: Yes, what would you recommend? \\
A: I'm happy to recommend the fish, It tastes delicious, and it is today's special. Our chef is from the coast, and loves seafood. Today's special is actually his favorite dish. so I'm sure it is a \\
Response: It does sound wonderful, maybe I'll try it .\\
\\
Generate the response following the following dialogue: + element\\

\textbf{For the multiple responses generation, CD-pred and CD-GT tasks, the prompt used is:}
Generate + str(k) + responses following this dialogue: + element\\
Number the generated sequences from 1 to + str(k) \\
Generated sequences: \\
1: 

\paragraph{Beluga PB}

In this case, 'element' still stands for the 3-turn context, and `labels' is the sequence of expected labels (\emph{e.g. `'inform, happiness'}). The expected labels can either come from the dataset (task CD-GT) or from the prediction of a BART generative model (task CD-pred). The prompt used is:
Generate the response following the given context : + element \\
The tone of the response must be + labels \\
Response: 

\newpage
\section*{Appendix C: Generation Example}
\label{sec:examples}
Let's consider the following dialogue:\\
\textbf{SPEAKER A: }Good morning. What's the matter with you?\\
\textbf{SPEAKER B: }Good morning, Doctor. I have a terrible headache.\\
\textbf{SPEAKER A: }All right, Young man. Tell me how it got started.\\
\\
The ground truth label is `inform'. The labels generated by the BART model are [`inform', `commissive']. The actual response to the dialogue is:\\
\textbf{ACTUAL RESPONSE: }Yesterday I had a runny nose. Now my nose is stuffed up. I have a sore throat. And I'm afraid I've got a temperature. I feel terrible.\\
The following table \ref{tab:examples_response} presents the answers generated by all the considered models and approaches.

{\renewcommand{\arraystretch}{1.2}%
\begin{table*}
  \centering
  \label{tab:dialogue}
  \begin{tabular} {p{50mm}|p{110mm}}
    \hline
    \textbf{Model} & \textbf{Response} \\
    \hline
    \textbf{GPT-2}             &     \\
    
    \hspace{3mm}\emph{GPT-2} 	      & It started on me at work and has continued on me at home. I am having a bad headache in the morning. I often sit down to eat at my desk but have to come home at night to recover. \\
    \hspace{3mm}\emph{GPT-2 CD-GT} 	    & It started when I was in school. My father knocked on my door asking if I wanted to come with him to the cafeteria to lunch. When I opened the door, I saw my father lying on the ground with a head wound. My father rushed to the scene to save him, But when I pressed him, I saw his bloody nose cut and swollen chin. My father was bleeding badly from the wound. \\
    \hspace{3mm}\emph{GPT-2 CD-pred} 	    & It started when I was in school. My father knocked on my door asking if I wanted to come with him to the cafeteria to lunch. When I opened the door, I saw my father lying on the ground with a head wound. My father rushed to the scene to save him, But when I pressed him, I saw his bloody nose cut and swollen chin. My father was bleeding badly from the wound.\\
    \hline  
    \hspace{3mm}\emph{GPT-2 Medium} 	      & I heard the news about the student strike and decided to go to the university administration. I asked them to give me a rest for a few days. But they didn't listen to me. I was under the impression that they would listen to me. But then I heard the news about the strike again. I went back to the university administration and again tried to get some help. \\
    \hspace{3mm}\emph{GPT-2 Medium CD-GT} 	    & Well, I just stood up too fast. I feel dizzy. \\
    \hspace{3mm}\emph{GPT-2 Medium CD-pred} 	    & Well, I just stood up too fast. I feel dizzy. \\
    \hline  

    \textbf{DialoGPT}            &      \\
    \hspace{3mm}\emph{DialoGPT small} 	  & It's a little complicated. First, It started out as a fever, And now it's a headache. \\
    \hspace{3mm}\emph{DialoGPT small CD-GT} 	& It's really a mystery. \\
    \hspace{3mm}\emph{DialoGPT small CD-GT} 	& It's really a mystery. \\
    \hline
    \hspace{3mm}\emph{DialoGPT medium} 	  & I guess I had a bad dream last night. \\
    \hspace{3mm}\emph{DialoGPT medium CD-GT} & Well, I was under the influence when I got up this morning. \\
    \hspace{3mm}\emph{DialoGPT medium CD-pred} & Well, I was under the influence when I got up this morning. \\
    \hline
    
    \textbf{BART}              & \\
    \hspace{3mm}\emph{BART base} 	    & What's the matter with you?\\
    \hspace{3mm}\emph{BART base CD-GT} 	  & I'm afraid you have a bad headache.\\
    \hspace{3mm}\emph{BART base CD-pred} 	  & I'm afraid you have a bad headache.\\
    \hline
    \hspace{3mm}\emph{BART large} 	    & Well, Young man, I'm afraid you'll have to stay in bed for three days. \\
    \hspace{3mm}\emph{BART large CD-GT} 	  & I see. Well, I'll prescribe some ointment and you'll be well soon.\\
    \hspace{3mm}\emph{BART large CD-GT} 	  & I see. Well, I'll prescribe some ointment and you'll be well soon.\\
    \hline
    
    \textbf{Beluga}             &     \\ 
    \hspace{3mm}\emph{Beluga} 	      & Yes, it feels like a sharp pain and it's really intense. It's like somebody is hammering inside my head. \\
    \hline
    \hspace{3mm}\emph{Beluga R CD-GT} 	  & Good morning, what's the matter with you? \\
    \hspace{3mm}\emph{Beluga R CD-pred} 	  & Good morning, what's the matter with you?\\
    \hline
    \hspace{3mm}\emph{Beluga PB CD-GT} 	  & Good morning, doctor. I have a terrible headache. The headache started late last night after I spent several hours studying. \\
    \hspace{3mm}\emph{Beluga PB CD-pred} 	  & What were you doing before you got the headache? \\
    \hline
  \end{tabular}
  \caption{Example of the response obtained by all the considered approaches to a same context sample.}
  \label{tab:examples_response}
\end{table*}

\newpage
\section*{Appendix D: Complete Form for Human Evaluation Step 3}
\label{sec:humaneval}

In this Appendix, we present the details and reference materials that were provided to the human judges during the evaluation task. Steps 1 and 2 relied on the definitions for Consistency and Specificity given in the paper. For Step 3, the annotators first had to tag each response with dialogue responses. Daily Dialog uses a system of 4 dialogue acts:

\begin{figure*}[!htbp] 
\centering
\label{fig:dialog}
\includegraphics[width=1\textwidth]{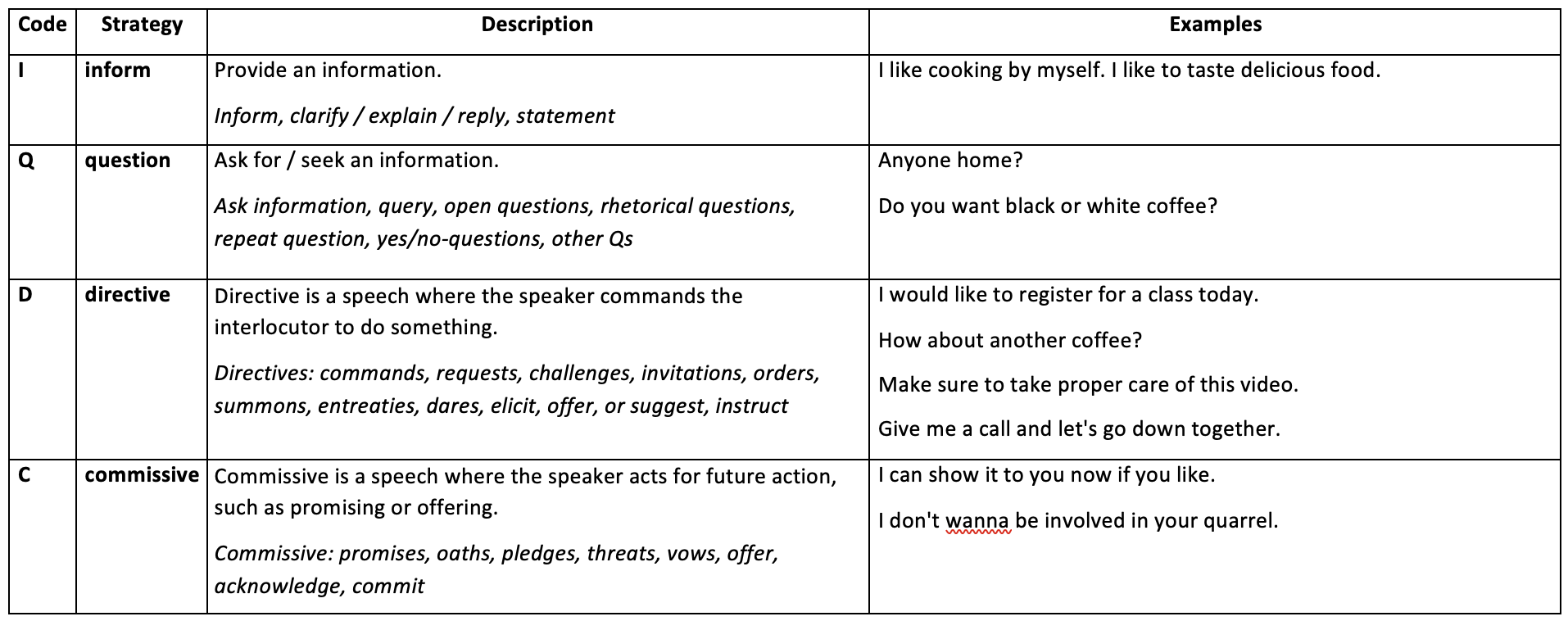}
\caption{4 dialogue acts used to annotate Daily Dialog, as well as some examples from the dataset to assist this task.}
\end{figure*}

For example, the response: I'm sorry to hear about Suzy's cold. Do you think you could ask someone from the family or close friends to help out? It might be best not to take her on the trip if she's not feeling well.

Will be tagged as: $<$I$>$ I'm sorry to hear about Suzy's cold.$<$/I$>$ $<$Q$>$ Do you think you could ask someone from the family or close friends to help out?$<$/Q$>$ $<$I$>$ It might be best not to take her on the trip if she's not feeling well.$<$/I$>$

Once the response is annotated with the dialogue acts, the judges must rate the following items:
\begin{figure*}[!htbp] 
\centering
\label{fig:socemocriteria}
\includegraphics[width=1\textwidth]{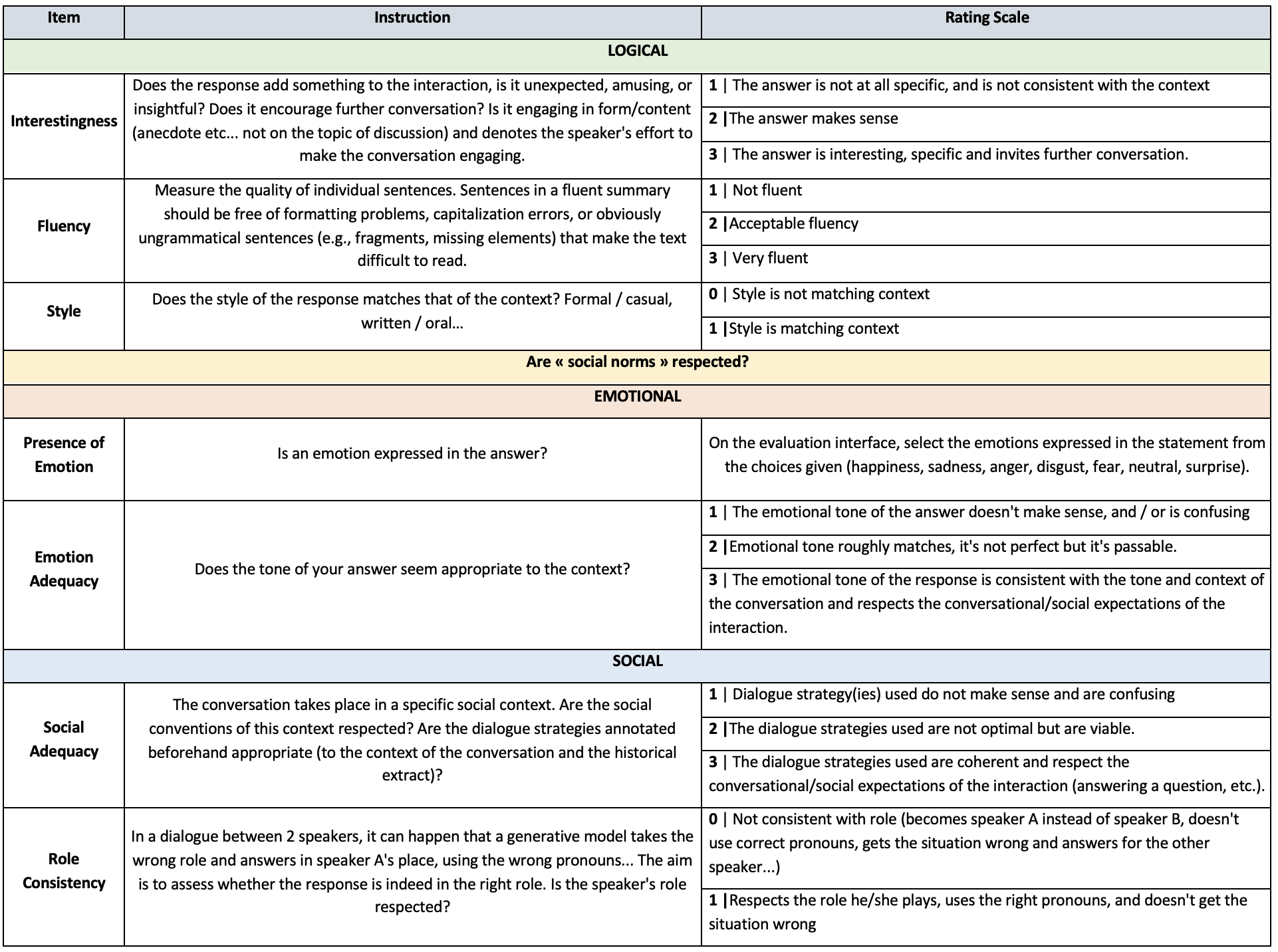}
\caption{Definition of each socio-emotional criteria rated in this evaluation, as well as the rating scale used for each item}
\end{figure*}

\newpage
\section*{Appendix E: Detailed Results of Human Evaluation}
\label{sec:humanevalscores}
In Table \ref{tab:results_conditioning_scores}, you will find the details of all scores obtained from the human evaluation we carried out on Daily Dialog. While the $socemo$ score is weighted by the number of responses by the model in the annotated sample, the logical, emotional and social ratings are unweighted. We weigh the fluency score similarly to the $socemo$ score to compare it to the Perplexity metric.

\renewcommand{\arraystretch}{1.2}%
\begin{table*}[!htbp]
    \centering
    \begin{small}
    \label{tab:dialogue}
    \begin{tabular} {p{40mm}|p{13mm}|p{13mm}|p{13mm}|p{13mm}|p{13mm}|p{13mm}|p{13mm}}
        \hline
        \textbf{Model} & \textbf{filtered} & \textbf{top3} & \textbf{socemo} & \textbf{logical} & \textbf{emotional} & \textbf{social} & \textbf{weighted fluency}\\ 
        \hline
        \textbf{GPT-2}                          &         &         &           &      &      & & \\
        \hspace{3mm}\emph{GPT-2 Small NO-CD} 	    & 33&9&13  & 90 & \textbf{100} & 94 &12 \\
        \hspace{3mm}\emph{GPT-2 Small CD-pred} 	    & 37&9&14 & 88 & 98 & 98& 13 \\
        \hspace{3mm}\emph{GPT-2 Small CD-GT} 	    & 37&9&14 & 88 & 98 & 98& 13\\
        \hline    
        \hspace{3mm}\emph{GPT-2 Medium NO-CD} 	    &\textbf{53} &21&28 & 91 & 99 & 98& 27 \\
        \hspace{3mm}\emph{GPT-2 Medium CD-pred} 	& \textbf{5}&19& 30 & 83 & 99 & 99 & 29 \\
        \hspace{3mm}\emph{GPT-2 Medium CD-GT} 	    & \textbf{5}&19& 30 & 93 & 99 & 99 & 29 \\
        \hline    
        \textbf{DialoGPT}                       &         &         &           &      &      &&\\
        \hspace{3mm}\emph{DialoGPT Small NO-CD}     & 37&11&18 & 90 & 98 & 98 & 17\\
        \hspace{3mm}\emph{DialoGPT Small CD-pred} 	& 4&13 &19 & 90 & 97& 98 & 18\\
        \hspace{3mm}\emph{DialoGPT Small CD-GT} 	& 4&13 &19 & 90 & 97 & 98 & 18 \\
        \hline
        \hspace{3mm}\emph{DialoGPT Medium NO-CD}    & \textbf{53}&16&15 & 89 & 99 & 99 & 14\\
        \hspace{3mm}\emph{DialoGPT Medium CD-pred}  & \textbf{52}&16&20 & 88 & \textbf{100} & \textbf{100} & 18\\        
        \hspace{3mm}\emph{DialoGPT Medium CD-GT}    & \textbf{52}&16&20 & 88 & \textbf{100} & \textbf{100}& 18\\
        \hline
        \textbf{BART}                           &         &         &           &      &      && \\
        \hspace{3mm}\emph{BART Base NO-CD} 	    & 32&7&13 & 87 & 97 & 95& 12\\
        \hspace{3mm}\emph{BART Base CD-pred} 	& 32&7&14 & 82  & 98 & 96 & 13\\
        \hspace{3mm}\emph{BART Base CD-GT} 	    & 32&7&14 & 82 & 98 & 96 & 13 \\
        \hline
        \hspace{3mm}\emph{BART Large NO-CD} 	& 42&9&19  & 89  & 99 & 99 & 19\\
        \hspace{3mm}\emph{BART Large CD-pred} 	& 45&12&19  & 88 & \textbf{100} & 99 & 19\\
        \hspace{3mm}\emph{BART Large CD-GT} 	& 45&12&19  &88 & \textbf{100} & 99 & 19 \\
        \hline
        \textbf{Beluga}                         &         &        &            &      &      & &\\ 
        \hspace{3mm}\emph{Beluga NO-CD} 	    & 42&25& 39 &93 & 98 &  \textbf{100} & 38\\
        \hspace{3mm}\emph{Beluga PB CD-pred} 	& \textbf{51}&\textbf{36} &44    & 93 & 98 & 98 & 44\\
        \hspace{3mm}\emph{Beluga PB CD-GT} 	    & 45&3 &\textbf{51} & 94 & 97 & 99 & 52\\
        \hline
        \textbf{Daily Dialog Reference} 	    & 97 &61 & 69  & 94 & 98 & \textbf{100} & 63\\
        \hline
    \end{tabular}
    \caption{All the results from the human evaluation: Step 1 - Filtering (column 1), Step 2 - top-3 (column 2) \& Step 3 socio-emotional annotation (column 3 is the global score, computed as the average of the three axes scores in columns 4-6).}
    \label{tab:results_conditioning_scores}
    \end{small}
\end{table*}

\section*{Appendix F: Results on New Daily Dialog Dataset}
\label{sec:resultsdd}
Instead of using the huggingface dataset, which was reported to have a significant overlap between the test and train sets, we use the splits provided in Daily Dialog's original paper, which do not display the same duplicate issue. In our original experiments, we had not fine-tuned our Beluga models (inference only), so those results are unaffected by the test-train set data overlap. We reran our code on the remaining models - BART, DialoGPT and GPT2 - using the same GPU and hyper-parameters as in the main paper). These results, available in Table \ref{tab:results_conditioning_F} are similar to those obtained with the test-train sets duplicates. While we do not claim that using the huggingface splits displaying duplicates did not have any negative impact on the training, this new set of results seems to indicate that this impact might not be too significant or invalidate the results shown in this study.
\begin{table*}[!htbp] 
    \centering
    \begin{small}
    \label{tab:dialogue}
    \begin{tabular} {p{36mm}|p{5mm}|p{5mm}|p{5mm}||p{6mm}|p{8mm}|p{7mm}|p{7mm}|p{12mm}|p{6mm}|}
        \hline
        \textbf{Model} & \textbf{Sacre bleu} & \textbf{Rouge} & \textbf{Bert score} & \textbf{CHRF} \\ 
        \hline 
        \emph{GPT-2 \textsubscript{b} NO-CD} & 96 & 12 & 86 & 13	    \\
        \emph{GPT-2 \textsubscript{b} CD-pred} \textsubscript{(R)}     	    & 103 & 12 &86 & 15  \\
        \emph{GPT-2 \textsubscript{b} CD-GT} \textsubscript{(R)}     	 & 103 & 12 &86 & 15  \\
        \hline   
        \emph{GPT-2 \textsubscript{M} NO-CD} & 176 & 14 & 87 & 14   \\
        \emph{GPT-2 \textsubscript{M} CD-pred} \textsubscript{(R)}     	 & 169 & 14 &87 & 16  \\
        \emph{GPT-2 \textsubscript{M} CD-GT} \textsubscript{(R)}    	     & 169 & 14 &87 & 16  \\
        \hline  \hline   
        \emph{DialoGPT \textsubscript{b} NO-CD}     & 99 & 13 & 86 & 12 \\
        \emph{DialoGPT \textsubscript{b} CD-pred} \textsubscript{(R)}     	 & 90 & 13 & 87 & 15  \\
        \emph{DialoGPT \textsubscript{b} CD-GT} \textsubscript{(R)}     & 90 & 13 & 87 & 15 \\
        \hline
        \emph{DialoGPT \textsubscript{M} NO-CD}     & 217 & 15 & 87 & 14  \\
        \emph{DialoGPT \textsubscript{M} CD-pred} \textsubscript{(R)}      & 233 & 16 & 87 & 17  \\   
        \emph{DialoGPT \textsubscript{M} CD-GT} \textsubscript{(R)}       & 233 & 16 & 87 & 17  \\
        \hline         \hline
        \emph{BART \textsubscript{b} NO-CD} 	    & 218 & 17 & 88 & 12  \\
        \emph{BART \textsubscript{b} CD-pred} \textsubscript{(R)}	 & 236 & 17 &87 & 18  \\
        \emph{BART \textsubscript{b} CD-GT} \textsubscript{(R)}	     & 236 & 17 &87 & 18  \\
        \hline
        \emph{BART \textsubscript{L} NO-CD} 	 & 303 & 18 & 87 & 14  \\
        \emph{BART \textsubscript{L} CD-pred} \textsubscript{(R)}	 & 356 & 19 &87 & 20  \\
        \emph{BART \textsubscript{L} CD-GT} \textsubscript{(R)}	 & 236 & 16 & 87 & 18  \\
        \hline
    \end{tabular}
    \caption{Comparative results of the experiments on conditioning response generation using multi-label sequences modelling social and emotional behaviours, on a different DailyDialog split, that does not feature any duplicate across the different sets. \\ Results are given in \%. \textsubscript{b} denotes the Base or Small model, \textsubscript{M} the Medium model and \textsubscript{L} the Large model.\\ \textsubscript{(R)} means the approach used is reranking, while \textsubscript{(Prompt)} refers to prompt-based generation.}
    \label{tab:results_conditioning_F}
    \end{small}
\end{table*}

\end{document}